\documentclass{article}

\usepackage{microtype}
\usepackage{graphicx}
\usepackage{caption,subfigure}
\usepackage{booktabs} 
\usepackage{enumerate}

\usepackage{amssymb,amsmath,amsthm}
\usepackage{mathrsfs}
\usepackage{array,multirow}
\usepackage{color}
\usepackage{mathtools}
\newtheorem{lemma}{Lemma}

\usepackage{relsize} 
\usepackage{xifthen}

\DeclarePairedDelimiterX{\norm}[1]{\lVert}{\rVert}{#1} 
\newcommand\abs[1]{\left|#1\right|}

\usepackage{tikz}

\newcommand{\Iset}     {\ensuremath{\mathcal{I}}}

\newcommand{\TENSORiset}     {\ensuremath{\mathcal{I}_{\text{TEN}}}}

\newcommand{\HCiset}     {\ensuremath{\mathcal{I}_{\text{HC}}^{D,\gamma}(R)}}

\newcommand{\ENHCweight}     {\ensuremath{\omega_{\text{ENHC}}^{D,\gamma,\zeta}(\kBS)}}
\newcommand{\ENHCiset}     {\ensuremath{\mathcal{I}_{\text{ENHC}}^{D,\gamma,\zeta}(R)}}

\newcommand{\aBS}      {\ensuremath{\boldsymbol{a}}}

\newcommand{\rBS}      {\ensuremath{\boldsymbol{r}}}

\newcommand{\kBS}      {\ensuremath{\boldsymbol{k}}}

\newcommand{\KBF}      {\ensuremath{\mathbf{K}}}

\newcommand{\TBS}      {\ensuremath{\boldsymbol{T}}}
\newcommand{\uBS}      {\ensuremath{\boldsymbol{u}}}

\newcommand{\xBS}      {\ensuremath{\boldsymbol{x}}}

\newcommand{\DeltaBS}      {\ensuremath{\boldsymbol{\Delta}}}
\newcommand{\gammaBS}      {\ensuremath{\boldsymbol{\gamma}}}
\newcommand{\omegaBS}      {\ensuremath{\boldsymbol{\omega}}}

\newcommand{\tauBS}      {\ensuremath{\boldsymbol{\tau}}}

\newcommand{\thetaBS}      {\ensuremath{\boldsymbol{\theta}}}
\newcommand{\XiBS}      {\ensuremath{\boldsymbol{\Xi}}}

\newcommand{\TORUSD}      {\ensuremath{\mathbb{T}^D}}

\newcommand{\fourierexpkdotx}      {\ensuremath{2\pi i \boldsymbol{k}\cdot\boldsymbol{x}}}

\newcommand{\ffourier}{\boldsymbol{\omega}} 

\usepackage{hyperref}

\usepackage[accepted]{icml2018}

\begin{document}

\twocolumn[
\icmltitle{Index Set Fourier Series Features for Approximating Multi-dimensional Periodic Kernels}



\icmlsetsymbol{equal}{*}

\begin{icmlauthorlist}
\icmlauthor{Anthony Tompkins}{ed}
\icmlauthor{Fabio Ramos}{ed}
\end{icmlauthorlist}

\icmlaffiliation{ed}{School of Information Technologies, The University of Sydney, New South Wales, 2006 Australia}

\icmlcorrespondingauthor{anthony.tompkins}{sydney.edu.au}

\icmlkeywords{periodic kernels, fourier series, random feature approximations, deterministic function approximation, bayesian linear regression}

\vskip 0.3in
]



\printAffiliationsAndNotice{}  

\begin{abstract}
Periodicity is often studied in timeseries modelling with autoregressive methods but is less popular in the kernel literature, particularly for higher dimensional problems such as in textures, crystallography, and quantum mechanics. Large datasets often make modelling periodicity untenable for otherwise powerful non-parametric methods like Gaussian Processes (GPs) which typically incur an $\mathcal{O}(N^3)$ computational burden and, consequently, are unable to scale to larger datasets. To this end we introduce a method termed \emph{Index Set Fourier Series Features} to tractably exploit multivariate Fourier series and efficiently decompose periodic kernels on higher-dimensional data into a series of basis functions. We show that our approximation produces significantly less predictive error than alternative approaches such as those based on random Fourier features and achieves better generalisation on regression problems with periodic data. 
\end{abstract}

\section{Introduction}
Although non-parametric methods \cite{gershman2012tutorial} are exceptionally flexible methods for statistical modelling, they inherently lack scalability. A particular non-parametric method based on covariance functions, a.k.a. \emph{kernels}, is the GP \cite{RasmussenBook} which has analytically tractable posterior and marginal likelihood. However, the inability to truly scale represents a major limitation for time series more general periodic problems which require computing and storing data-dependent covariance matrices which have the potential to grow as one acquires more data. While there have been various efforts to approximate GPs with lower rank solutions based on \emph{inducing points} \cite{snel2006ghar, hensman2013gaussian} these methods are still constrained by their data-dependence. 

To alleviate data-dependence, recent innovations take advantage of spectral decompositions of the kernel functions with explicit data-\emph{independent} features. Motivated in the key works of \cite{RahimiRecht2007,rahimi2009weighted} and termed \emph{Random Fourier Features} (RFFs) is the idea of explicit data-\emph{independent} feature maps using ideas from harmonic analysis and sketching theory. By approximating kernels these maps allow scalable inference with simple linear models. Various approximations to different kernels have followed and include polynomial kernels \cite{pham2013fast,pennington2015spherical}, dot-product kernels \cite{kar2012random}, histogram and $\gamma$-homogenous kernels \cite{li2010random, vedaldi2012efficient}, and approximations based upon the so-called \emph{triple-spin} \cite{choromanski2016triplespin}: \cite{le2013fastfood, yang2015carte, felix2016orthogonal}. A recent work of note, quasi-Monte Carlo features (QMC) \cite{sindhwani2014high}, uses deterministic sequences on the hypercube to approximate shift invariant kernels. In particular, we compare this body of work with the Halton sequence which was found to be highly performant, and with the generalised Halton using the permutation in \cite{rainville2012evolutionary}.

While much work has been done for data-independent kernel approximations such as RFFs, as opposed to Nystr\"{o}em \cite{williams2001using, gittens2013revisiting}, there is limited work on such approximations of \emph{periodic} kernels. Two recent works \cite{SolinSarka2014, tompk2018fourier} explore approximations for periodic kernels in univariate timeseries where some response varies periodically with respect to time - however it is not clear how to tractably generalise such decompositions into multiple dimensions where the response varies periodically as a function of \emph{multiple} inputs. In this paper, leveraging deterministic feature space decompositions of multivariate periodic kernels, we demonstrate the efficacy of our solution in terms of both kernel gram matrix reconstruction and generalisation error on synthetic and real data.

Specifically, in our contributions we provide: i) a generalisation of Fourier series approximations of stationary periodic kernels into the multivariate domain using sparse index sets; ii) an efficient sparse decomposition for multivariate Fourier series representations of periodic kernels; iii) an extension to non-isotropic periodic kernels; iv) a general bound for the cardinality of the resulting full and sparse feature sets; v) an upper bound to the truncation error for the multivariate feature approximation; and vi) a numerically stable implementation for the periodic Squared Exponential kernel (in supplementary material).

We finally compare in detail the proposed method against recent state-of-the-art kernel approximations in terms of the kernel approximation error as well as the predictive downstream error. Empirical results on real datasets further demonstrate that deterministic index set based features provide significantly improved convergence generalisation properties by reducing both the data samples \emph{and} the number of features required for equivalently accurate predictions. 

\section{Preliminaries}
We begin by reviewing the relevant kernel literature. The inference setup is deferred to the supplementary.

\textbf{Notation}. Let $\mathbb{R}$ represent the set of real numbers, $\mathbb{Z}$ the set of integers, $\mathbb{Z}^+_0$ the set of non-negative integers, and $\mathbb{N}^+$ the set of positive integers. For any arbitrary set $\mathbb{Y} \neq \emptyset$, let $\mathbb{Y}^D$ be its Cartesian product $\mathbb{Y} \times ... \times \mathbb{Y}$ repeated $D$ many times where $D \geq 1, D \in \mathbb{N}$. Let $\mathbb{T}^D := [a,b]^D$ represent the $D$-dimensional torus, or the circle with $D=1$. Throughout this paper, $D$ represents the spatial dimension and $R \in \mathbb{Z}^+_0$ is the maximum \emph{refinement}. The refinement may be interpreted as the multidimensional analogue for the integer valued univariate Fourier series truncation $K$.  
\subsection{Fourier Features for Approximating Kernels}
\textbf{Random Fourier Features.}
The result of note from RFFs is Bochner's Theorem \cite{Bochner1933} which allows one to approximate a shift invariant kernel as $\kappa(\xBS, \xBS^{\prime}) \approx  \langle\hat{\Phi}(\xBS),\hat{\Phi}(\xBS^{\prime})\rangle_{\mathbb{C}^C}$. It states that a complex-valued function $g: \mathbb{R}^D \rightarrow \mathbb{C}$ is positive definite if and only if it is the Fourier transform of a finite non-negative Borel measure $\mu$ on $\mathbb{R}^D$: $g(\xBS) = \hat{\mu}(\xBS) = \int_{\mathbb{R}^D}e^{-i\xBS^T\ffourier}d\mu (\ffourier),\quad \forall \xBS \in \mathbb{R}^{D}$. Without loss of generality, assuming measure $\mu$ has associated scaled probability density function $p$, we have:
\begin{equation}
\kappa(\xBS,\xBS^{\prime}) \approx \frac{1}{C}\sum^C_{j=1}e^{-i(\xBS-\xBS^{\prime})^T\omegaBS_C} \label{eq:fourier_full_complex} = \langle\hat{\Phi}(\xBS),\hat{\Phi}(\xBS^{\prime})\rangle_{\mathbb{C}^C},
\end{equation}
where the kernel approximation is determined by \eqref{eq:fourier_full_complex}. $C$ is the number of spectral samples from the density $p(\cdot)$. This is in fact an MC approximation to the integral. In equation \eqref{eq:fourier_full_complex} and using the relation $e^{-i\tauBS\cdot\omegaBS_k} = \cos(\omega\cdot\tauBS) - i\sin(\omega\cdot\tauBS)$ and the fact that real kernels have no imaginary part, we can exploit the trigonometric identity:
\begin{equation}
\cos(u-v)=\cos(u)\cos(v)+\sin(u)\sin(v).
\label{cos_ident_sum}
\end{equation}
giving the $2C$-dimensional mapping $\hat{\Phi} : \mathcal{X} \rightarrow \mathbb{R}^{2C}$:
\begin{equation}
\hat{\Phi}({\xBS}) = \frac{1}{\sqrt{C}}\Big[\cos(\xBS\omegaBS_c^T), \sin(\xBS\omegaBS_c^T)\Big]_{c}^{C}.
\label{RFF_FEAT}
\end{equation}
\textbf{Fourier Series Features.} A method of approximating periodic equivalents of standard shift invariant kernels is depicted in \cite{tompk2018fourier}, where the authors define a feature based framework based on Fourier series expansions of periodic kernel functions. The key idea is that the Fourier series of a periodic kernel can be harmonically decomposed in a deterministic manner. Given a Fourier series representation of some time-domain kernel $\kappa(t)$, the truncated Fourier series is:
\begin{equation}
\kappa(t) \approx F_{k}[\kappa(t)]=\sum_{k=-\infty}^{\infty} c_{k} e^{ik\omega_{0}t},
\label{FSF_eqn}
\end{equation}
with Fourier coefficients $c_0 = \frac{1}{2L}\int_{-L}^{L}f(t)dt$ and $c_k = \frac{1}{2L}\int_{-L}^{L}f(t)e^{-ik\omega_{0}t} dt,\quad\forall k \in \mathbb{N}^+$. Here, $L$ is the half period and $\omega_0$ is the fundamental frequency. By solving for $c_k$ one can determine the corresponding analytic Fourier series coefficients. While the above work is tractable in the univariate case, we show in this work it does not straightforwardly expand to higher dimensions due to computationally intractable complexity in the number of approximating coefficients. This limitation for high dimensional periodicity motivates the main contribution of this work where we show there are sparse multivariate expansions for periodic kernels.

\subsection{Fourier Series in Higher Dimensions}
Our goal is to represent high dimensional periodic kernels. In the primal space of the full kernel, such a composition can be represented as a product of $D$-many independent periodic kernels on each $d=1,...,D$ dimension. It is known that product compositions in the primal space of the kernel have an equivalent cartesian product operation in the dual (feature) space. The key idea is that by noting particular results from harmonic theory on weighted subspaces of the Wiener algebra \cite{bonsall2012complete,kammerer2014phd, kammerer2015approximation}, we can use sparse sampling grids to efficiently represent multivariate periodic kernels. 

Consider a sufficiently smooth multivariate periodic function $f:\mathbb{T}^D \rightarrow \mathbb{C}, D \in \mathbb{N}^+$. A function $f$ can formally be represented as its multivariate Fourier series:
\begin{equation}
f(\xBS) = \sum_{\kBS \in \mathbb{Z}^D} \hat{f}_{\kBS} e^{\fourierexpkdotx},
\end{equation}
with its Fourier series coefficients  $\hat{f}_{\kBS}, \kBS \in \mathbb{Z}^D$ defined as $\hat{f}_{\kBS} := \int_{\mathbb{T}^D} f(\xBS)e^{\fourierexpkdotx} d\xBS$. Let $\Pi_\Iset$ denote the space of all multivariate trigonometric polynomials supported on some arbitrary \emph{index set} $\Iset \subset \mathbb{Z}^D$. This is simply a finite set of integer vectors. Denote the \emph{cardinality} of the set $\Iset$ as $\abs{\Iset}$. Practically, we are interested in a good approximating Fourier partial sum $S_{\Iset}f \in \Pi_{\Iset}$ supported on some suitable index set $\Iset$. We can now define weighted function spaces of the Wiener algebra: $\mathcal{A}_{\omega}(\TORUSD) := \{f \in L_1(\TORUSD): \sum_{\kBS\in \mathbb{Z}^D} \hat{f}_{\kBS}e^{\fourierexpkdotx}, \sum_{\kBS \in \mathbb{Z}^D} \omega(\kBS)\abs{\hat{f}_{\kBS}} < \infty \}$ where we define a \emph{weight function} $\omega : \mathbb{Z}^D \rightarrow [1, \infty]$, which attempts to characterise the decay of Fourier coefficients $\hat{f}_{\kBS}$ of all functions $f \in \mathbb{A}_{\omega}(\TORUSD)$ such that $\hat{f}_{\kBS}$ decreases faster than the weight function in terms of $\kBS$. That is to say that the decay of the Fourier coefficients defines the smoothness of the function $f$ we which are trying to approximate with a partial sum. Furthermore, subspaces of the Wiener algebra contain functions of specific types of smoothness; specifically, we investigate index sets $\Iset$ on \emph{weighted} $l_p$\emph{-balls} (LPB), $0 < p \leq \infty$, which correspond to functions of isotropic smoothness, and also the \emph{Weighted Energy Norm Hyperbolic Cross} (ENHC), which is a generalisation of the Weighted Hyperbolic Cross (HC), for functions of dominating mixed smoothness.

\subsection{Index Sets}
While the complete expansion of a univariate Fourier series appears plausible, it is clear that for some maximum isotropic order $R$ for the coefficients that the cardinality $\lvert \mathcal{I} \rvert = R^D$ of the complete multivariate Fourier series expansion is computationally intractable; this cardinality corresponds to the tensor product of multiple 1-dimensional Fourier series for each dimension. Making things worse, the computational burden is amplified when we consider the expanded representation required for the separable feature decomposition when the products of harmonic terms $\varrho$ of the Fourier series themselves must be expanded into sums of cosines. All these considered, it would thus be desirable to; i) maintain high function approximation accuracy; and ii) minimise the total number of coefficients. 

To this end we introduce the next ingredient required to approximate multivariate kernels: \emph{index sets}. We introduce a variety of $D$-dimensional weighted index sets $\mathcal{I}_D$ from the multivariate function approximation literature with their formal definitions. Here, we briefly review the ENHC and refer the interested reader to the supplementary material for extended coverage of additional index sets. The index set definitions to follow refer to the supporting frequency lattice for multivariate Fourier series $f : \mathbb{T}^D \rightarrow \mathbb{C}$ where $f(\xBS) = \sum_{\mathbf{k}\in \mathbb{Z}^D} \hat{f}_{\mathbf{k}}e^{2\pi i \mathbf{k} \cdot \xBS}$ and $\hat{f}_{\mathbf{k}}$ are the Fourier series coefficients.

\textbf{Weighted Energy Norm Hyperbolic Cross} \label{subsec_ENHC}
\begin{figure}[t]
\vskip 0.1in
\begin{center}
\centerline{\includegraphics[width=\columnwidth]{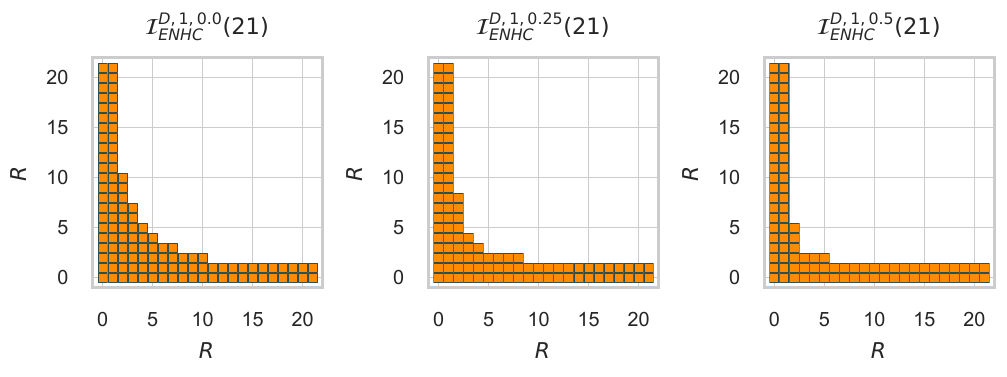}}
\caption{ENHC index set for various $\zeta$.}
\label{fig:E0140_enhc_2D}
\end{center}
\vskip -0.1in
\end{figure}

This section introduces the \emph{Weighted Energy Norm Hyperbolic Cross} (ENHC) index set \cite{dung2016hyperbolic}, which is a generalisation of the Hyperbolic Cross (HC) index set \cite{zaremba1972methode,hallatschek1992fouriertransformation, kammerer2013reconstructing}. HC based index sets are a more suitable method for approximating functions in spaces of dominating mixed smoothness, a.k.a. \emph{Korobov spaces}  \citep{gnewuch2014weighted}. In the case of the explicit tensor product of multivariate approximations (see supplementary regarding $l_p$-ball), the cardinality grows exponentially fast in  dimension $D$ and is computationally intractable for accurately representing kernels in dimensions $D\gtrapprox 5$. The ENHC enables approximations of functions with a large proportion of the Fourier coefficient mass centered at the origin. 

The weight function we consider for the ENHC is:
\begin{equation}
\ENHCweight = \max(1,\norm{\kBS}_1)^{\frac{\zeta}{\zeta-1}} \prod_{d=1}^{D} \max(1, \gamma_{d}^{-1}\abs{k_d})^{\frac{1}{1-\zeta}}
\label{weight_function_enhc}
\end{equation}
with dimension-dependent weighting parameter $\gammaBS = (\gamma_d)^{\infty}_{d=1} \in [0,1]^{\mathbb{N}^+}$ and sparsity parameter $\zeta \in [0,1)$. We can now define the Weighted Energy Norm Hyperbolic Cross index set of refinement $R$:
\begin{equation}
\ENHCiset := \{ \kBS \in \mathbb{Z}^D : \ENHCweight \leq R \}.
\end{equation}
The primary difference between the ENHC and the HC index sets is that the cardinality of the ENHC for the case $\zeta \in (0,1)$, which interestingly has a cardinality that isn't bounded by the term $(\log R)^{D-1}$ which occurs in the HC (see supplementary for details). We show in corresponding experiments how the performance of the ENHC in higher dimensions is able to closely resemble that of the Tensor or Total order index sets and continue to perform effectively for higher dimensions where the $l_p$-ball becomes computationally intractable.

\section{Index Set Fourier Series Features}
The goal of our work is to show how multivariate Fourier series representations of kernels with sparse approximation lattices allow efficient and deterministic feature decompositions of multivariate periodic kernels. Formally, we define the shift invariant multivariate periodic kernel approximation as a multivariate Fourier series expansion supported on an arbitrary index set $\Iset$:
\begin{equation}
\begin{split}
\kappa_{\text{per}}(\xBS,\xBS^{\prime}) &\approx \sum_{\kBS \in \Iset} \hat{f}_{\kBS} e^{2\pi i \boldsymbol{k}\cdot(\xBS-\xBS^{\prime})} \label{eq:ISFSF_full} \\
										&\approx \langle\hat{\Phi}(\xBS),\hat{\Phi}(\xBS^{\prime})\rangle_{\mathbb{C}^C},
\end{split}
\end{equation}
for some explicit feature map $\hat{\Phi}$ and multivariate Fourier series coefficients $\hat{f}_{\kBS}$.

We begin by explaining how one obtains Fourier series coefficients for univariate Fourier series kernel approximations. We follow with our feature construction which we term Index Set Fourier Series Features (ISFSF). We then introduce a sparse construction reducing total feature count for no loss of accuracy. We show it is straightforward to represent non-isotropic hyperparameters and perform non-isotropic approximations. Finally, we give an upper bound for the multivariate truncation error as well as an interesting numerically ideal asymptotic of the Bessel-to-exponential term in the Fourier coefficient term of the periodic Squared Exponential kernel (provided in supplementary).

\subsection{Univariate Periodic Kernel Approximations}
We demonstsrate first how one constructs a stationary periodic kernel and its corresponding Fourier series decomposition. It is possible to construct a periodic kernel from any stationary kernel by applying the \emph{warping} \cite{MacKayBook} $\uBS(x)=[\cos(x),\sin(x)]$ to data $x$ and then passing the warped result into any standard stationary kernel. Specifically, by performing the warping to a stationary kernel with the general squared distance metric $\norm{x - x^\prime}^2$ and replacing $x$ with $\uBS(x)$ we have:
\begin{align}
\begin{split}
& \norm{\uBS(x)-\uBS(x^{\prime})}^2 \\
 	& \qquad = (\sin(x)-\sin(x^{\prime}))^2 + (\cos(x)-\cos(x^{\prime}))^2 \\
	& \qquad = 4\sin^2\Big(\frac{x-x^{\prime}}{2}\Big)
	= 2(1-\cos(x-x^{\prime})),
	\label{eq:rbf_per_dist}
\end{split}
\end{align}
To demonstrate the warping process in full, as an example, consider the well known Squared Exponential (SE) kernel  \cite{RasmussenBook} $\kappa_{\text{SE}}({\xBS - \xBS^\prime}) =  \exp\Big(-\frac{\norm{\xBS - \xBS^\prime}^2}{2 l^2}\Big)$ with lengthscale $l$. After performing \eqref{eq:rbf_per_dist} we recover the periodic SE kernel: $\kappa_{\text{perSE}}(x,x^{\prime}) = \exp\Big( - \frac{\cos(w_0 \tau)-1}{l^2}\Big),$ where $\tau = x - x^{\prime}$. Solving the coefficients for \eqref{FSF_eqn} yields the univariate solution for a Fourier series representation $\kappa(\tau)\approx F_{k}[\kappa{\tau}] = \sum^{\infty}_{k=-\infty}q_k^2 \cos(k\omega_0 \tau)$. For the periodic SE 
\begin{equation}
q^2_{k} = 
     \begin{cases}
       \frac{I_{k}(l^{-2})}{\exp(l^{-2})} &\quad\text{if}\ k=0,\\
       \frac{2I_{k}(l^{-2})}{\exp(l^{-2})} &\quad\text{if}\ k=1,2,...,K,\\
     \end{cases}
\end{equation}
where $K$ is the truncation factor of the partial sum of the Fourier series \cite{tompk2018fourier}. We use this univariate $q_k$ on a per-dimension basis for our multivariate feature construction.
\subsection{ISFSF Feature Construction} \label{isfsf_section}
This section presents our main contribution for approximating high dimensional periodic kernels. We show that simply using multivariate Fourier series is not sufficient to represent a stationary kernel in a decomposable way. Using \emph{index sets} for multivariate Fourier series we present a feature construction involving recursion of a trigonometric identity and the realisation there is a computationally efficient way to calculate the required sign coefficients within the harmonic feature terms using a cartesian combination. We also show that the resulting feature set is sparse and can thus be further reduced with no loss of accuracy. 

We introduce the method with an illustrative example for the bivariate case of $D=2$, and follow with the general construction for all $D \in \mathbb{N}^+$. An extended worked example for $D=\{2,3\}$ is provided in the supplementary. Let us now consider the bivariate case of $D=2$ on $[0,1]^2$:
\begin{equation}
\kappa(\DeltaBS) \approx \sum_{r_1,r_2=0}^{R-1} q_{r_1,r_2}\cos(a_{r_1}^{(1)}\Delta^{(1)})\cos(a_{r_2}^{(2)}\Delta^{(2)}),
\end{equation}
where $\DeltaBS = \xBS - \xBS^{\prime} =  \big[\Delta^{(1)},\Delta^{(2)},..,\Delta^{(D)}\big]$ and $a$ is some dimension and refinement index dependent constant. $\rBS=[r_1,r_2]$ is a coordinate index vector subscripted by per-dimension refinement index $r_d$ from some index set with max refinement $R=[R_1,R_2]$. $R_1 = R_2$ for the isotropic case of equal weights $\gammaBS = [\gamma_1,\gamma_2]$ with $\gamma_1 = \gamma_2$. Multi-dimensional Fourier series coefficients $q_{r_1,r_2}$ are evaluated at index $(r_1, r_2)$ where the shorthand notation refers to the per-dimension product of the respective coefficients: $q_{r_1,r_2} = q^{(1)}_{r_1}q^{(2)}_{r_2}$. For our feature expansion we are interested in the data-dependent trigonometric term of the form  $\cos(\Delta^{(1)})\cos(\Delta^{(2)})$, where $\Delta^{(1)}$ and $\Delta^{(2)}$ are the data-dependent differences for dimension 1 and 2 respectively. In order for this to be decomposed into the form of \eqref{cos_ident_sum} we require an additional trigonometric identity:
\begin{equation}
\cos(u)\cos(v) = \frac{1}{2}[\cos(u-v) + \cos(u+v)].
\label{cos_ident_product}
\end{equation}
For the bivariate case, a single application of \eqref{cos_ident_product} admits the solution. To explicitly illustrate this, we set refinement $R=2$ and use the full tensor grid index set $\TENSORiset(R)$. The tensor grid results in the given index set:
\begin{equation*}
\TENSORiset(2) = 
\begin{bmatrix}
\Iset_0\\
\Iset_1\\
\Iset_2\\
\Iset_3\\
\end{bmatrix} =
\begin{bmatrix}
0&0\\
0&1\\
1&0\\
1&1\\
\end{bmatrix}.
\end{equation*} 
which, after applying \eqref{cos_ident_product}, allows a straightforward decomposition into separable features. E.g. for $\Iset_0$ and $\Iset_1$:
\begin{align*}
\kappa(\Delta&) \approx\\
&\quad   \frac{q_{0,0}}{2}\big(\cos(0\Delta^{(1)} - 0\Delta^{(2)}) + \cos(0\Delta^{(1)} + 0\Delta^{(2)})\big)\\
				   &+ \frac{q_{0,1}}{2}\big(\cos(0\Delta^{(1)} - 1\Delta^{(2)}) + \cos(0\Delta^{(1)} + 1\Delta^{(2)})\big).
\end{align*}
This trigonometric term can now be decomposed using \eqref{cos_ident_sum} because we have independent cosine terms; that is to say a product term  $\cos(\Delta^{(1)})\cos(\Delta^{(2)})$ has a decomposition as sums of harmonic terms (cosines) where the inner terms involve the elementwise product $\aBS \odot \XiBS \odot \Delta$ where $\XiBS$ is the sign matrix. Continuing the above example, this sign coefficient matrix is: $\XiBS = 
\begin{bmatrix}
+1&-1\\
+1&+1\\
\end{bmatrix}.$
Having shown the expanded trigonometric form, we can now write the general form for the harmonic monomial product expansion for any particular $i^{th}$ index set coordinate $\Iset_{i}$ for some arbitrary index set $\Iset$ from the partial sum for a $D$ dimensional multivariate Fourier series:
\begin{equation}
\begin{split}
\varrho(\Iset_i)  &= \prod_{d=1}^D q_{i}^{(d)}\cos(a_{i}^{(d)}\Delta^{(d)}) \\
 &= \prod_{d=1}^D q_i^{(d)}\cos(i^{(d)} \omega_0^{(d)}(x^{(d)}-x^{\prime(d)})),
\end{split}
\label{isfsf_product_expansion}
\end{equation}
where $\omega_0^{(d)}$ is the $d^{\text{th}}$ dimension's fundamental frequency. The term \eqref{isfsf_product_expansion}, after repeated application of \eqref{cos_ident_product}, admits the following decomposable form:
\begin{equation}
\begin{split}
\varrho(\Iset_i)  =   \frac{1}{2^{(D-1)}} \sum_{j=0}^{2^{D-1}} \rho\cos\big( \boldsymbol{a} \odot \boldsymbol{\Xi}_j \odot \boldsymbol{\Delta} \big),
\end{split}
\label{isfsf_sum_expansion_naive}
\end{equation}
with
\begin{align}
&\begin{aligned}
\rho &= \prod_{d=1}^{D} q_{i}^{(d)},
\end{aligned} \\
&\begin{aligned}
\boldsymbol{a} &= \big[a_{i}^{(1)}, a_{i}^{(2)},...,a_{i}^{(D)}\big], \\
               &= \big[i^{(1)}\omega_0^{(1)},i^{(2)}\omega_0^{(2)},...,i^{(D)}\omega_0^{(D)}\big],
\end{aligned} \\               
&\begin{aligned}
\boldsymbol{\Xi} &= \big[+\mathbf{1}^{\frown}  (+1, -1)^{(D-1)} \big], 
\end{aligned}\\
&\begin{aligned}
\boldsymbol{\Delta} &= \big[\Delta^{(1)},\Delta^{(2)},..,\Delta^{(D)}\big], \\
                    &=\big[x^{(1)} - x^{\prime(1)},x^{(2)} - x^{\prime(2)},...,x^{(D)} - x^{\prime(D)}\big],
\end{aligned}    
\label{isfsf_sum_terms}                
\end{align}
where $(+1,-1)^{(D-1)}$ is the $(D-1)$-times Cartesian combination of the ordered integer set. $(+1,-1)$, $v^{\frown}\Lambda$ denotes a horizontal concatenation between some matrix $v$ of length $\abs{\Lambda}$ and every element in some ordered set $\Lambda$, and $\odot$ refers to the Hadamard (element-wise) product. Using \eqref{cos_ident_sum} the complete decomposed feature is thus given as:
\begin{equation}
\hat{\Phi}_{\Iset}^{\text{full}}({\xBS}) = \sqrt{\frac{\boldsymbol{\rho}}{2^{(D-1)}}} \Big[  \cos\big( \boldsymbol{a} \odot \boldsymbol{\Xi}_j \odot \xBS \big), \sin\big( \boldsymbol{a} \odot \boldsymbol{\Xi}_j \odot \xBS \big)   \Big].
\label{isfsf_full_feature_representation}
\end{equation}
We can now state the cardinality $\abs{\hat{\Phi}_{\Iset}^{\text{full}}(\xBS)}$ (i.e. resulting dimensionality) of the resulting \emph{decomposed} feature map $\hat{\Phi}_{\Iset}^{\text{full}}(\xBS)$ over some index set $\Iset$.
%
%
\begin{lemma}Cardinality of Fourier series feature map for some arbitrary index set $\Iset(R)$. Let $C_{\Iset} = \abs{\Iset(R)}$ be the cardinality of some given index set of refinement $R$, and let the dimension $D\in\mathbb{N}^+$ be given. Let $C_{\hat{\Phi}}^{\text{full}}=\abs{\hat{\Phi}_{\Iset}^{\text{full}}(\xBS)}$ denote the cardinality of the decomposed feature The following holds (see supplementary for proof): 
\begin{equation*}
\abs{\Iset(R)} \leq \abs{\hat{\Phi}_{\Iset}^{\text{full}}(\xBS)} = {C_{\Iset}}2^{D}.
\end{equation*}
\end{lemma}
\subsection{Sparse Construction} 
Although suitable, the decomposable form for the multivariate Fourier series features can be improved. An ideal feature representation should not just approximate our kernel \emph{well} but should do it \emph{efficiently}. For ISFSF, this involves the data-dependent term $\cos(\cdot)$, the occurrences of which we want to minimise. Scrutinising the product form in \eqref{isfsf_product_expansion} a little closer, notice the term $i^{(d)}$ is an integer $i \in \mathbb{Z}^+_0$ which clearly contains the value $0$. This means that for \emph{all} refinement coordinates at the $0^{\text{th}}$ refinement for any dimension we have $\cos(0 \cdot) = 1$, therefore the "feature" is simply $1$ times some data-\emph{independent} coefficient. Furthermore, since any \emph{single} $\cos(\cdot)$ term in the product \eqref{isfsf_product_expansion} contributes to a multiplier of $2$ features before exponentiation due to the repeat application of \eqref{cos_ident_product}, we don't unnecessarily want to include features that will simply evaluate to a constant, e.g. if we have $q^{(1)}\cos(u)q^{(2)}\cos(0) = q^{(1)}q^{(2)}\cos(u)$ and $q^{(1)}\cos(u)q^{(2)}\cos(0)q^{(3)}\cos(0)q^{(4)}\cos(x)=q^{(1)}q^{(2)}q^{(3)}q^{(4)}\cos(u)\cos(x)$ giving 2 and 4 final features respectively. Masking out specific product terms where $i=0$ and retaining only the data-independent coefficients $q_i$ in \eqref{isfsf_product_expansion}, we admit a sparse construction. We now define a \emph{masking} function $\varkappa$ over some function $g(i)$ with $i \in \mathbb{Z}^+_0$:
\begin{equation}
\varkappa\Big(g(i)\Big) = 
     \begin{cases}
       1 &\quad \text{for } i=0,\\
       g(i) &\quad \text{otherwise}.\\
     \end{cases}
\end{equation}
Now we can define the augmented product expansion:
\begin{equation}
\varrho(\Iset_i)  = \prod_{d=1}^D q_i^{(d)} \varkappa\Big(    \cos(i^{(d)} \omega_0^{(d)}(x^{(d)}-x^{\prime(d)}))    \Big), \\
\label{isfsf_product_expansion_nozero}
\end{equation}
where $\omega_0^{(d)}$ is the $d^{\text{th}}$ dimension's fundamental frequency. The term \eqref{isfsf_product_expansion_nozero}, after repeated application of \eqref{cos_ident_product}, admits the following decomposable form: $\varrho(\Iset_i)  =   \frac{1}{2^{(D-1)}} \sum_{j=0}^{2^{D-1}} \rho   \varkappa\Big(   \cos\big( \boldsymbol{a} \odot \boldsymbol{\Xi}_j \odot \boldsymbol{\Delta} \big)    \Big)$ with all corresponding terms identical to \eqref{isfsf_sum_terms}.
We are now ready to state the sparse feature decomposition:
\begin{equation}
\begin{split}
\hat{\Phi}_{\Iset}^{\text{mask}}({\xBS}) = \sqrt{\frac{\boldsymbol{\rho}}{2^{(D-1)}}} \Big[&\varkappa\Big(   \cos\big( \boldsymbol{a} \odot \boldsymbol{\Xi}_j \odot \xBS \big)    \Big),\\
&\varkappa\Big(   \sin\big( \boldsymbol{a} \odot \boldsymbol{\Xi}_j \odot \xBS \big)    \Big)\Big].
\label{isfsf_masked_feature_representation}
\end{split}
\end{equation}
We now give an improved cardinality for the decomposed sparse ISFSF feature map. We emphasise this improved feature map cardinality is for the \emph{same reconstructing} accuracy. To determine the cardinality of this sparse feature map, let:
\begin{equation}
\eta(\Iset_i) = \sum_{d=1}^{D}{\big[ \Iset_i \neq 0\big]},
\end{equation} 
define a function that counts for a particular coordinate $\Iset_i$ the non-zero indexes. Essentially, we want to count which $\cos(\cdot)$ terms to keep, and these will always occur at coordinates with per-dimension refinement not equal to 0. 
\begin{lemma}Cardinality of sparse Fourier series feature map for some arbitrary index set $\Iset(R)$. Let $C_{\Iset} = \abs{\Iset(R)}$ be the cardinality of some given index set of refinement $R$, and let the dimension $D\in\mathbb{N}^+$ be given. Let $C_{\hat{\Phi}}^{\text{mask}}=\abs{\hat{\Phi}_{\Iset}^{\text{mask}}(\xBS)}$ then denote the cardinality of the decomposed index set Fourier series feature. The following holds (see supplementary for proof):
\begin{equation*}
\abs{\Iset(R)} \leq \abs{\hat{\Phi}_{\Iset}^{\text{mask}}(\xBS)} = \sum^{\abs{\Iset}}_{i=1} 2^{\eta(\Iset_i)} \leq \abs{\hat{\Phi}_{\Iset}^{\text{full}}(\xBS)}  \leq {C_{\Iset}}2^{D}.
\end{equation*}
\end{lemma}
To appreciate this improvement in real terms with an example, imagine we have an HC index set $\HCiset$ with $D=5$, refinement $R=10$ and $\gammaBS=\boldsymbol{1}$. This index set has cardinality $C_{\Iset}=1432$ giving a full ISFSF feature expansion of cardinality $C_{\Phi}^{full}=45824$ whereas the sparse feature map has significantly reduced cardinality of $C_{\Phi}^{mask}=16893$ for the same approximating accuracy. 

\subsection{Multivariate Truncation Error} \label{subsec_truncation_error}
\begin{figure}[t]
\vskip 0.1in
\begin{center}
\centerline{\includegraphics[width=\columnwidth]{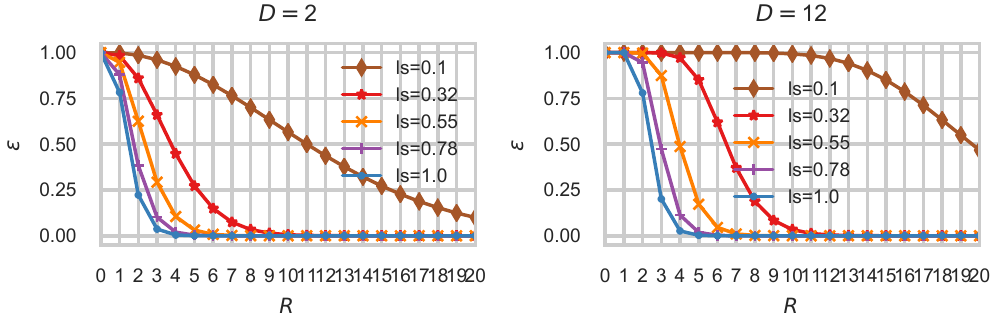}}
\caption{Visualisation of the multivariate truncation error, for the periodic SE kernel, for linearly increasing refinements $R=[0,20]$, dimensions $D=\{2,12\}$ and isotropic lengthscales $l=[0.1,1.0]$.}
\label{fig:E0153_isfsf_trunc_error}
\end{center}
\vskip -0.1in
\end{figure}
We have from the univariate truncation error \cite{SolinSarka2014} for some $k \in \mathbb{N}^+$ that $\abs{\cos(\omega_0 k \tau)} \leq 1$, and furthermore $\sum_{k=0}^{\infty}q^2_{k} = 1$, which is to say the sum of the coefficients converges to 1. It is straightforward to extend this to the multivariate case by considering the tensor index set product expansion. We have then: $\prod^{D}_{d=1}\abs{\cos(\omega_0^{(d)} r^{(d)} \tau^{(d)})} \leq 1$. Additionally, since  $\max(\kappa(\tauBS)) = 1$ we define the multivariate truncation error: 
\begin{equation}
\epsilon(R, l) = 1- \prod^{D}_{d=1}\Big[ \sum_{r=0}^{R-1}q_{r}^{(d)2}\Big], \label{mv_approx_error}
\end{equation} 
where $R$ is the refinement, superscript $(d)$ refers to some evaluation on the $d^{\text{th}}$ dimension, $l$ is the kernel lengthscale. $q_r^{(d)}$ refers to the approximated kernel's Fourier coefficient at refinement index $r$ with the $d^{\text{th}}$ dimension's corresponding lengthscale.

\subsection{Non-Isotropy}
\textbf{Non-isotropic Hyperparameters}. Directly analogous to hyperparamter Automatic Relevance Determination (ARD) in GPs \cite{mackay1996bayesian,neal2012bayesian} we demonstrate how, with ISFSF, one can straightforwardly model non-isotropic kernel hyperparameters such as lengthscale and period. There are two considerations. 

Firstly, in the case of non-isotropic periodicities, we simply have the modification of the scalar $T$ into a vector $\TBS = [T^{(d=1)},...,T^{(d=D)}]$ resulting in the vector $\omegaBS_{0}= \frac{2\pi}{\TBS}$.

Secondly for non-isotropic hyperparameters such as the \emph{lengthscale}, we again have a modification of a scalar value to a vector. For some arbitrary scalar hyperparameter $\theta$, the non-isotropic extension is simply the vector $\thetaBS = [\theta^{(d=1)},...,\theta^{(d=D)}]$. 
\begin{figure}[t]
\vskip 0.1in
\begin{center}
\centerline{\includegraphics[width=\columnwidth]{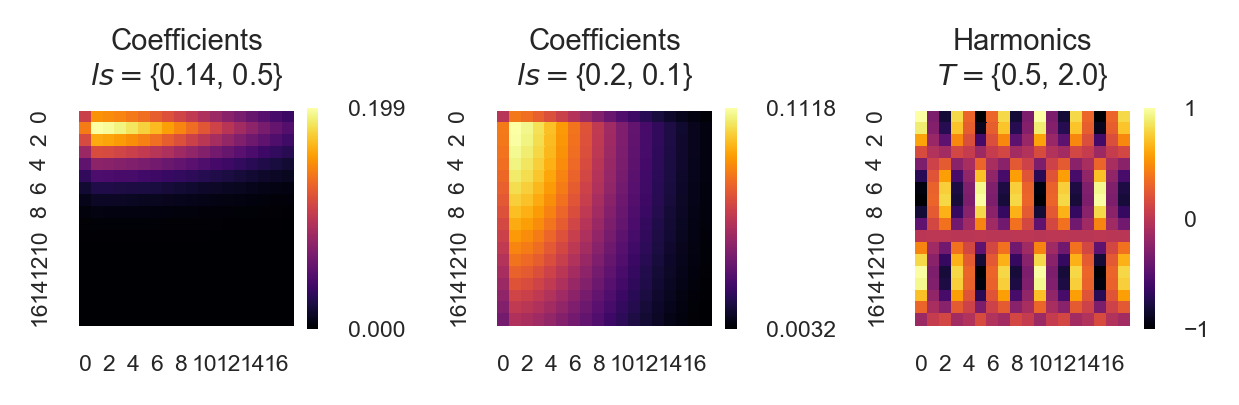}}
\caption{Visualisation of the Fourier coefficients and harmonic terms for a 2D approximation with max refinement $R=18$. From left to right, the first two images depict non-isotropic lengthscales and the third image represents data-dependent harmonic terms with non-isotropic periodicity when $\tau = \xBS - \xBS^{\prime} = 0.25$.}
\label{fig:E0151_2d_noniso_hypers}
\end{center}
\vskip -0.1in
\end{figure}
Figure \ref{fig:E0151_2d_noniso_hypers} demonstrates more clearly how non-isotropic lengthscale and periodicity affect the multivariate coefficients and data-dependent harmonic terms. Observe the lower lengthscales correspond to slower decay in the coefficients reflecting the hyperparameter-dependent upper bound given in section \ref{subsec_truncation_error}.

\textbf{Non-isotropic Approximation}. We now briefly remark upon non-isotropic approximation. Recall from Figure \ref{fig:E0153_isfsf_trunc_error} the smaller the lengthscale the more refinements required for a particular approximation accuracy, and conversely, for larger lengthscales fewer refinements. This motivates the idea that for any particular multivariate kernel being approximated with Fourier series, it is not necessary to have isotropic refinements for all dimensions. We achieve no more than some improvement $\epsilon_{\Delta}$ in approximating error between two levels of refinement $\epsilon_{\Delta} = \epsilon_{A} - \epsilon_{B}$, where $A$ and $B$ refer to refinement levels. Formally, non-isotropic approximation is defined by the refinement weight parameter $\gammaBS = (\gamma_d)^{\infty}_{d=1} \in [0,1]^{\mathbb{N}^+}$ for each weighted index set.
\section{Experiments}
\begin{table*}[ht]
\small
\centering
\begin{tabular}{llllllllllllll}
\toprule
                    &       & \multicolumn{5}{c}{RMSE} &\hspace{1em} & \multicolumn{5}{c}{MNLL}\\ 
\toprule
                    &       & \multicolumn{5}{c}{Total components} & &  \multicolumn{5}{c}{Total components}\\                     
\toprule
Dataset       & Method & $49$  & $93$   & $201$ & $397$ & $793$  & & $49$  & $93$  & $201$ & $397$ & $793$ \\ 
\midrule
\multirow{4}{*}{Pores}   & ENHC   & \textbf{0.0787}  &\textbf{0.0701}  & \textbf{0.0612} & \textbf{0.0608} & \textbf{0.0608} & & \textbf{0.3971} &\textbf{-0.9284} &\textbf{-1.3578} & \textbf{-1.3504}   & \textbf{-1.2833}	 \\
                         & RFF+W  &0.271    &0.2615  &0.2329  & 0.131  & 0.0614 & & 9.7526 &4.3514  &1.1269  & -0.5845   & -1.1675 \\
                         & HAL+W  &0.2737   & 0.258  &0.2267  & 0.1271 & 0.0612 & & 9.7804 &4.2501  & 1.0625 & -0.6224   & -1.1665 \\
                         & GHAL+W & 0.265   &0.2436  & 0.2063 & 0.1316 & 0.0611 & & 9.3598 & 3.8556 & 0.773  & -0.5759   & -1.1653 \\
                         \midrule
\multirow{4}{*}{Rubber}   & ENHC   & \textbf{0.1119}  &\textbf{0.1042}   & \textbf{0.1013}  & \textbf{0.1026}  & \textbf{0.1024} & &\textbf{10.1399}   & 3.5833  & \textbf{0.554}  & \textbf{-0.363}    &-0.7092\\
                         & RFF+W   &0.1491   &0.1436   &0.1443   & 0.1363  & 0.1102 & &16.3536   & 7.4633  &2.4248  & 0.3837    &-0.6986\\
                         & HAL+W   &0.1523   &0.1412  &0.1465    & 0.1319  &0.1111  & &  16.8435 & \textbf{2.3266}  & 7.5562 & 0.2451    & -0.7092\\
                         & GHAL+W  &0.146    &0.1421  &0.1344    &  0.1244 &  0.107 & &15.8572   & 7.1348  &2.1058  &   0.0976  & \textbf{-0.7331} \\
                         \midrule
\multirow{4}{*}{Tread}  & ENHC    & 0.0614  & 0.0617 & \textbf{0.0622} & 0.0624 & 0.0623 & & \textbf{-0.6146}& \textbf{-1.0925} &\textbf{-1.2306} &  \textbf{-1.2716}  & \textbf{-1.2747 }\\
                         & RFF+W  & \textbf{0.0611}  & \textbf{0.0615} & 0.0625  & 0.0622 & 0.0618 & & -0.5593& -1.0157& -1.1273  &  -1.2301  & -1.176\\
                         & HAL+W  &0.0613   & 0.0619 & 0.0625 & 0.0633 & 0.0626 & & -0.5874& -1.0001& -1.1205&  -1.2236  & -1.2115\\
                         & GHAL+W &0.0622   & 0.0623 &0.0628  & \textbf{0.0621} & \textbf{0.0617} & & -0.5293& -1.1558&-1.2033 &  -0.9971 & -1.2376 \\
\bottomrule
\end{tabular}
\caption{Comparison of predictive RMSE and MNLL with our method using the ENHC alongside Fourier Feature methods, for increasing number of components. The number of training and test points in each dataset are $N_{\text{train}}=12675$ and $N_{\text{test}}=4225$ respectively. Note FF methods by construction produce an even number of total components and so have an equivalent number of ISFSF components $+1$.}
\label{texture_extrap}
\end{table*}
We make three critical assessments of the proposed periodic kernel approximation. First, we directly compare the approximation with the full kernel in terms of Gram matrix reconstruction; second we will perform a comparison of predictive qualities against an analytic periodic function (see supplementary material); and finally we directly compare our features to the corresponding state of the art approximations on large scale real world textured image data.
\subsection{Quality of Kernel Approximation}
\begin{figure}[t]
\captionsetup[subfigure]{aboveskip=2pt,belowskip=2pt}
\centering
\begin{subfigure}  \centering
\includegraphics[width=\columnwidth]{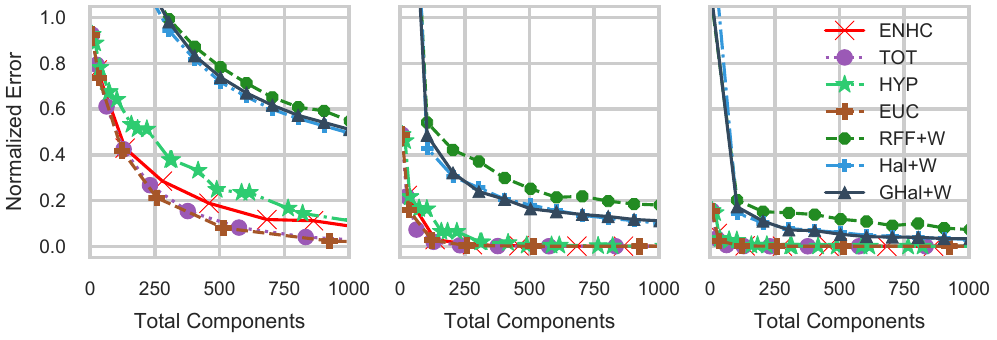}
\label{fig:E0158_3D_all_comparison}
\end{subfigure}
\begin{subfigure}  \centering
\includegraphics[width=\columnwidth]{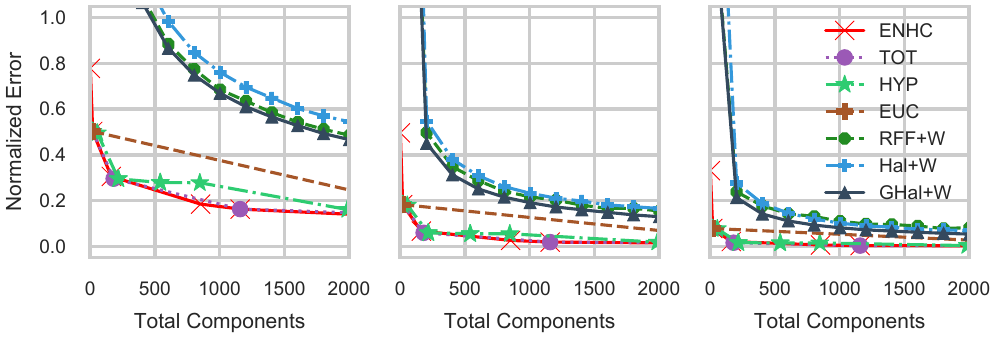}
\label{fig:E0158_9D_all_comparison}
\end{subfigure}
\caption{Comparison of normalised Frobenius error between Fourier Feature methods (RFF, Hal, GHal) with periodic warpings, and Index Set Fourier Series with various index sets. \textbf{Top Row}: $D=3$, $ls=\{0.5, 1.0, 1.5\}$. \textbf{Bottom Row}: $D=9, ls=\{1.5, 2.0, 2.5\}$. Note the ENHC uses a weighting of $\gamma=\frac{2}{3}$ for $D=9$. See supplementary extended comparisons.}
\label{fig:E0158_frobenius_all_comparison}
\end{figure}
We first analyse the proposed feature in terms of the reconstruction error between a true Gram matrix $\KBF$, using the analytic periodic kernel, and its approximated Gram $\tilde{\KBF}$ where each $\tilde{\KBF}_{i,j}=\hat{\kappa}(x_i, y_j)$. For all comparisons the metric we use is the Normalized Frobenius error $\frac{\|\tilde{\mathbf{K}}-\mathbf{K}\|_{F}}{\|\mathbf{K}\|_{F}}$ using $N=4000$ uniformly drawn samples drawn on a single periodic interval $[-2,2]$. The primary comparison in Figure \ref{fig:E0158_frobenius_all_comparison} compares the effects of various index set constructions, RFFs, QMC (Halton, Generalised Halton), and the following index sets: Energy Norm Hyperbolic Cross (ENHC), Total Order (TOT), Hyperbolic (HYP), and Euclidean (EUC). The supplementary contains an extended comparison including comparisons of index set parameters, additional dimensions, and nuances of the reconstruction.

The first observation we can make from Figure \ref{fig:E0158_frobenius_all_comparison} is that for lower dimensions $D\approx 3$ the best performing features are those with the Euclidean degree or Total order index sets which correspond to LPBall index sets with parameter $p$ set to 2 and 1 respectively. Of the index sets the Hyperbolic and Energy Norm Hyperbolic Cross perform the worst in particular for smaller lengthscales. Overall the index sets all perform significantly better than the warped Fourier Feature methods, amongst which, the original MC based RFF performs the worst and the standard Halton sequence appears to perform marginally better than the generalised Halton. 

As the number of dimensions increases the Total and Euclidean index sets become intractable due to their heavy dependency on cross-dimensional products of data-dependent harmonic terms. Considering FF methods, the approximation accuracy of the standard Halton sequence falls behind even RFFs while the generalised Halton remains consistently ahead. As we have seen, as the dimensionality increases, the Total and Euclidean index sets have no parameterisation that allows them to scale properly; indeed their flexibility is due entirely being a specific instance of the LPBall index set which can give sparser and sparser \emph{Hyperbolic} index sets. On the other hand, the ENHC can be parameterised by a sparsity $\zeta$ and weighting factor $\gamma$ giving additional flexibility.

\subsection{Generalisation Error}
We evaluate predictive performance with Root Mean Square Error (RMSE) and Mean Negative Log Loss (MNLL). The MNLL allows us to account for the model's predictive mean and predictive uncertainty. It is defined as $\text{MNLL} = \frac{1}{N} \sum^{N}_{i=1}\frac{1}{2}\log(2\pi\sigma^{*}_i)) + \frac{(\mu^{*}_i - f^{*}_i)^2}{2\sigma^{*}_i}$ where $\sigma^{*}_i$, $\mu^{*}_i$, and $f^{*}_i$ are respectively the predictive standard deviation, predictive mean, and true value at the $i^{\text{th}}$ test point. This section demonstrates the generalisation performance of a single multidimensional periodic kernel on image texture data. We use images from \cite{wilson2014fast} with the same $12675$ train and $4225$ test set pixel locations $\xBS \in \mathbb{R}^2$.
\begin{figure}[t]
\vskip 0.1in
\begin{center}
\centerline{\includegraphics[width=\columnwidth]{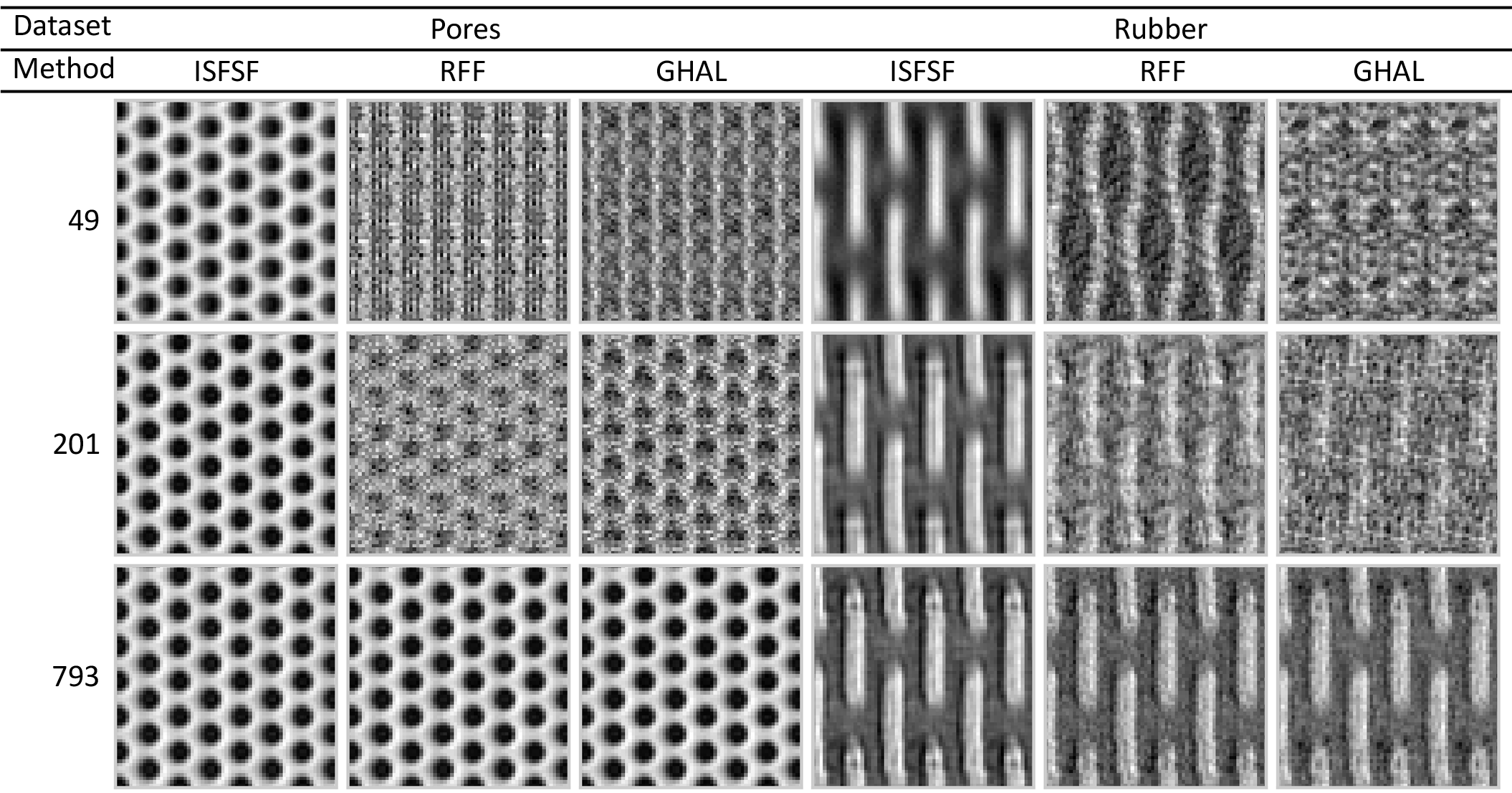}}
\caption{Visualisation of the predicted missing area for the \emph{Pores} and \emph{Rubber} datasets. From left to right, each column represents predictions made using ISFSF, RFF, and GHAL. From top to bottom, each row represents an increasing number of features used at $49, 201, 793$ respectively.}
\label{fig:E159_pores_visual_comparison}
\end{center}
\vskip -0.1in
\end{figure}
Overall, it can be seen that ISFSF based feature provide significantly improved generalisation capability in terms of reconstructing the periodic missing data. For exactly the same kernel, both qualitatively and quantitatively the results show clear advantages over the alternative Fourier Feature methods. Results may be seen in both Table \ref{texture_extrap} and Figure \ref{fig:E159_pores_visual_comparison}.

\textbf{Pores}. In this dataset we can see the RMSE and MNLL for the ISFSF based features (using the ENHC) perform the best in all cases. The RMSE performs exceedingly well even with only 49 features almost equaling the performance of RFF and QMC methods which require 794 features. In both RMSE and MNLL, the ISFSF with 201 features outperforms FF based methods using 794 features.

\textbf{Rubber}. For this image depicting the pattern of a rubber math, in each group of feature counts, ISFSF outperforms all other methods and is again nearly equal in performance with 49 features to the FF methods using 794 features. The generalised Halton marginally outperforms all methods in the case of 794 features, and the Halton at 94 features. Figure \ref{fig:icml_comparison_rubber_SUPPLEMENTARY} show the predicted regions, for the \emph{rubber} dataset, for various feature types and increasing number of feature components.

\textbf{Tread}. The last dataset we provides detail comparison is a textured metal tread pattern (see supplementary material). The resulting performance of this dataset is interesting because for all features, the RMSE performance is approximately equal for all feature counts. Additionally, while the RMSE performance remains relatively stable, the MNLL favours the ISFSF in all cases, with further improvements the larger the feature count.

\section{Discussion}
The ability to model periodicity is a unique problem in machine learning. Modelling with kernels, in particular with the GP, introduces problems related to data-dependent model complexity. We show that in the multivariate domain this dependence may be lifted by taking advantage of harmonic decompositions of kernel functions allowing scalable inference. Crucially, we introduce effective sparse approximation to multivariate periodic kernels using multivariate Fourier series with sparse index set based sampling grids allowing efficient feature space periodic kernel decompositions. We thus demonstrate the ability of even a single ISFSF based feature to effectively generalise, in terms of both RMSE and MNLL, with few features. 

There are several interesting problems to further investigate regarding ISFSF based decompositions, including i) generalised learning of kernels, before an ISFSF decomposition, that are dependent on the data and assume no a-priori structure \cite{wilson2013gaussian, jang2017scalable}; and arbitrary index sets that make no immediate assumption on the decay of the Fourier coefficients of the learned kernel \cite{ermakov1975monte,kammerer2014phd}.
\bibliography{icml_references}

\begin{thebibliography}{37}
\providecommand{\natexlab}[1]{#1}
\providecommand{\url}[1]{\texttt{#1}}
\expandafter\ifx\csname urlstyle\endcsname\relax
  \providecommand{\doi}[1]{doi: #1}\else
  \providecommand{\doi}{doi: \begingroup \urlstyle{rm}\Url}\fi

\bibitem[Bochner(1933)]{Bochner1933}
Bochner, S.
\newblock Monotone funktionen, stieltjessche integrale und harmonische analyse.
\newblock \emph{Mathematische Annalen}, 108:\penalty0 378--410, 1933.
\newblock URL \url{http://eudml.org/doc/159644}.

\bibitem[Bonsall \& Duncan(2012)Bonsall and Duncan]{bonsall2012complete}
Bonsall, Frank~F and Duncan, John.
\newblock \emph{Complete normed algebras}, volume~80.
\newblock Springer Science \& Business Media, 2012.

\bibitem[Choromanski et~al.(2016)Choromanski, Fagan, Gouy-Pailler, Morvan,
  Sarlos, and Atif]{choromanski2016triplespin}
Choromanski, Krzysztof, Fagan, Francois, Gouy-Pailler, Cedric, Morvan, Anne,
  Sarlos, Tamas, and Atif, Jamal.
\newblock Triplespin-a generic compact paradigm for fast machine learning
  computations.
\newblock \emph{arXiv preprint arXiv:1605.09046}, 2016.

\bibitem[D{\~u}ng et~al.(2016)D{\~u}ng, Temlyakov, and
  Ullrich]{dung2016hyperbolic}
D{\~u}ng, Dinh, Temlyakov, Vladimir~N, and Ullrich, Tino.
\newblock Hyperbolic cross approximation.
\newblock \emph{arXiv preprint arXiv:1601.03978}, 2016.

\bibitem[Ermakov(1975)]{ermakov1975monte}
Ermakov, Serge{\u \i}~Mikha{\u \i}lovich.
\newblock \emph{Die Monte-Carlo-Methode und verwandte Fragen}, volume~72.
\newblock Oldenburg, 1975.

\bibitem[Felix et~al.(2016)Felix, Suresh, Choromanski, Holtmann-Rice, and
  Kumar]{felix2016orthogonal}
Felix, X~Yu, Suresh, Ananda~Theertha, Choromanski, Krzysztof~M, Holtmann-Rice,
  Daniel~N, and Kumar, Sanjiv.
\newblock Orthogonal random features.
\newblock In \emph{Advances in Neural Information Processing Systems}, pp.\
  1975--1983, 2016.

\bibitem[Gershman \& Blei(2012)Gershman and Blei]{gershman2012tutorial}
Gershman, Samuel~J and Blei, David~M.
\newblock A tutorial on bayesian nonparametric models.
\newblock \emph{Journal of Mathematical Psychology}, 56\penalty0 (1):\penalty0
  1--12, 2012.

\bibitem[Gittens \& Mahoney(2013)Gittens and Mahoney]{gittens2013revisiting}
Gittens, Alex and Mahoney, Michael~W.
\newblock Revisiting the nystr{\"o}m method for improved large-scale machine
  learning.
\newblock \emph{J. Mach. Learn. Res}, 28\penalty0 (3):\penalty0 567--575, 2013.

\bibitem[Gnewuch et~al.(2014)Gnewuch, Mayer, and Ritter]{gnewuch2014weighted}
Gnewuch, Michael, Mayer, Sebastian, and Ritter, Klaus.
\newblock On weighted hilbert spaces and integration of functions of infinitely
  many variables.
\newblock \emph{Journal of Complexity}, 30\penalty0 (2):\penalty0 29--47, 2014.

\bibitem[Hallatschek(1992)]{hallatschek1992fouriertransformation}
Hallatschek, Klaus.
\newblock Fouriertransformation auf d{\"u}nnen gittern mit hierarchischen
  basen.
\newblock \emph{Numerische Mathematik}, 63\penalty0 (1):\penalty0 83--97, 1992.

\bibitem[Hensman et~al.(2013)Hensman, Fusi, and Lawrence]{hensman2013gaussian}
Hensman, James, Fusi, Nicolo, and Lawrence, Neil~D.
\newblock Gaussian processes for big data.
\newblock In \emph{Uncertainty in Artificial Intelligence}, pp.\  282.
  Citeseer, 2013.

\bibitem[Jang et~al.(2017)Jang, Loeb, Davidow, and Wilson]{jang2017scalable}
Jang, Phillip~A, Loeb, Andrew, Davidow, Matthew, and Wilson, Andrew~G.
\newblock Scalable levy process priors for spectral kernel learning.
\newblock In \emph{Advances in Neural Information Processing Systems}, pp.\
  3943--3952, 2017.

\bibitem[K{\"a}mmerer(2013)]{kammerer2013reconstructing}
K{\"a}mmerer, Lutz.
\newblock Reconstructing hyperbolic cross trigonometric polynomials by sampling
  along rank-1 lattices.
\newblock \emph{SIAM Journal on Numerical Analysis}, 51\penalty0 (5):\penalty0
  2773--2796, 2013.

\bibitem[K{\"a}mmerer(2014)]{kammerer2014phd}
K{\"a}mmerer, Lutz.
\newblock \emph{High dimensional fast Fourier transform based on rank-1 lattice
  sampling}.
\newblock PhD thesis, Technische Universit{\"a}t Chemnitz, 2014.

\bibitem[K{\"a}mmerer et~al.(2015)K{\"a}mmerer, Potts, and
  Volkmer]{kammerer2015approximation}
K{\"a}mmerer, Lutz, Potts, Daniel, and Volkmer, Toni.
\newblock Approximation of multivariate periodic functions by trigonometric
  polynomials based on rank-1 lattice sampling.
\newblock \emph{Journal of Complexity}, 31\penalty0 (4):\penalty0 543--576,
  2015.

\bibitem[Kar \& Karnick(2012)Kar and Karnick]{kar2012random}
Kar, Purushottam and Karnick, Harish.
\newblock Random feature maps for dot product kernels.
\newblock In \emph{International conference on artificial intelligence and
  statistics}, pp.\  583--591, 2012.

\bibitem[Le et~al.(2013)Le, Sarl{\'o}s, and Smola]{le2013fastfood}
Le, Quoc, Sarl{\'o}s, Tam{\'a}s, and Smola, Alex.
\newblock Fastfood-approximating kernel expansions in loglinear time.
\newblock In \emph{Proceedings of the international conference on machine
  learning}, volume~85, 2013.

\bibitem[Li et~al.(2010)Li, Ionescu, and Sminchisescu]{li2010random}
Li, Fuxin, Ionescu, Catalin, and Sminchisescu, Cristian.
\newblock Random fourier approximations for skewed multiplicative histogram
  kernels.
\newblock In \emph{Joint Pattern Recognition Symposium}, pp.\  262--271.
  Springer, 2010.

\bibitem[MacKay(1996)]{mackay1996bayesian}
MacKay, David~JC.
\newblock Bayesian methods for backpropagation networks.
\newblock In \emph{Models of neural networks III}, pp.\  211--254. Springer,
  1996.

\bibitem[MacKay(1998)]{MacKayBook}
MacKay, David~JC.
\newblock Introduction to {G}aussian processes.
\newblock \emph{NATO ASI Series F Computer and Systems Sciences}, 168:\penalty0
  133--166, 1998.

\bibitem[Neal(2012)]{neal2012bayesian}
Neal, Radford~M.
\newblock \emph{Bayesian learning for neural networks}, volume 118.
\newblock Springer Science \& Business Media, 2012.

\bibitem[Pennington et~al.(2015)Pennington, Felix, and
  Kumar]{pennington2015spherical}
Pennington, Jeffrey, Felix, X~Yu, and Kumar, Sanjiv.
\newblock Spherical random features for polynomial kernels.
\newblock In \emph{Advances in Neural Information Processing Systems}, pp.\
  1846--1854, 2015.

\bibitem[Pham \& Pagh(2013)Pham and Pagh]{pham2013fast}
Pham, Ninh and Pagh, Rasmus.
\newblock Fast and scalable polynomial kernels via explicit feature maps.
\newblock In \emph{Proceedings of the 19th ACM SIGKDD international conference
  on Knowledge discovery and data mining}, pp.\  239--247. ACM, 2013.

\bibitem[Rahimi \& Recht(2007)Rahimi and Recht]{RahimiRecht2007}
Rahimi, Ali and Recht, Ben.
\newblock Random features for large-scale kernel machines.
\newblock \emph{Conference on Neural Information Processing Systems}, 2007.

\bibitem[Rahimi \& Recht(2009)Rahimi and Recht]{rahimi2009weighted}
Rahimi, Ali and Recht, Benjamin.
\newblock Weighted sums of random kitchen sinks: Replacing minimization with
  randomization in learning.
\newblock In \emph{Advances in neural information processing systems}, pp.\
  1313--1320, 2009.

\bibitem[Rainville et~al.(2012)Rainville, Gagn{\'e}, Teytaud, Laurendeau,
  et~al.]{rainville2012evolutionary}
Rainville, De, Gagn{\'e}, Christian, Teytaud, Olivier, Laurendeau, Denis,
  et~al.
\newblock Evolutionary optimization of low-discrepancy sequences.
\newblock \emph{ACM Transactions on Modeling and Computer Simulation (TOMACS)},
  22\penalty0 (2):\penalty0 9, 2012.

\bibitem[Rasmussen \& Williams(2006)Rasmussen and Williams]{RasmussenBook}
Rasmussen, Carl~Edward and Williams, Christopher K.~I.
\newblock \emph{{G}aussian {P}rocesses for {M}achine {L}earning}.
\newblock MIT Press, 2006.

\bibitem[Sindhwani \& Avron(2014)Sindhwani and Avron]{sindhwani2014high}
Sindhwani, Vikas and Avron, Haim.
\newblock High-performance kernel machines with implicit distributed
  optimization and randomization.
\newblock \emph{arXiv preprint arXiv:1409.0940}, 2014.

\bibitem[Snelson \& Ghahramani(2006)Snelson and Ghahramani]{snel2006ghar}
Snelson, E. and Ghahramani, Z.
\newblock Sparse gaussian processes using pseudo-inputs.
\newblock In \emph{Advance in Neural Information Processing Systems},
  volume~18, pp.\  1257, 2006.

\bibitem[Solin \& S{\"a}rkk{\"a}(2014)Solin and S{\"a}rkk{\"a}]{SolinSarka2014}
Solin, Arno and S{\"a}rkk{\"a}, Simo.
\newblock Explicit link between periodic covariance functions and state space
  models.
\newblock In \emph{Proceedings of the Seventeenth International Conference on
  Artificial Intelligence and Statistics}, volume~33, pp.\  904--912, 2014.

\bibitem[Tompkins \& Ramos(2018)Tompkins and Ramos]{tompk2018fourier}
Tompkins, Anthony and Ramos, Fabio.
\newblock Fourier feature approximations for periodic kernels in time-series
  modelling.
\newblock In \emph{AAAI Conference on Artificial Intelligence}, 2018.

\bibitem[Vedaldi \& Zisserman(2012)Vedaldi and Zisserman]{vedaldi2012efficient}
Vedaldi, Andrea and Zisserman, Andrew.
\newblock Efficient additive kernels via explicit feature maps.
\newblock \emph{IEEE transactions on pattern analysis and machine
  intelligence}, 34\penalty0 (3):\penalty0 480--492, 2012.

\bibitem[Williams \& Seeger(2001)Williams and Seeger]{williams2001using}
Williams, Christopher~KI and Seeger, Matthias.
\newblock Using the nystr{\"o}m method to speed up kernel machines.
\newblock In \emph{Advances in neural information processing systems}, pp.\
  682--688, 2001.

\bibitem[Wilson \& Adams(2013)Wilson and Adams]{wilson2013gaussian}
Wilson, Andrew and Adams, Ryan.
\newblock Gaussian process kernels for pattern discovery and extrapolation.
\newblock In \emph{International Conference on Machine Learning}, pp.\
  1067--1075, 2013.

\bibitem[Wilson et~al.(2014)Wilson, Gilboa, Nehorai, and
  Cunningham]{wilson2014fast}
Wilson, Andrew~G, Gilboa, Elad, Nehorai, Arye, and Cunningham, John~P.
\newblock Fast kernel learning for multidimensional pattern extrapolation.
\newblock In \emph{Advances in Neural Information Processing Systems}, pp.\
  3626--3634, 2014.

\bibitem[Yang et~al.(2015)Yang, Wilson, Smola, and Song]{yang2015carte}
Yang, Zichao, Wilson, Andrew, Smola, Alex, and Song, Le.
\newblock A la carte--learning fast kernels.
\newblock In \emph{Artificial Intelligence and Statistics}, pp.\  1098--1106,
  2015.

\bibitem[Zaremba(1972)]{zaremba1972methode}
Zaremba, SK.
\newblock La m{\'e}thode des ``bons treillis'' pour le calcul des
  int{\'e}grales multiples.
\newblock In \emph{Applications of number theory to numerical analysis}, pp.\
  39--119. Elsevier, 1972.

\end{thebibliography}
\bibliographystyle{icml2018}
\end{document}